\documentclass{article}
\usepackage{ijcai20}

\usepackage{times}

\usepackage{soul}
\usepackage{url}
\usepackage[hidelinks]{hyperref}
\usepackage[utf8]{inputenc}
\usepackage[small]{caption}
\usepackage{graphicx}
\usepackage{amsmath}
\usepackage{booktabs}
\title{Identifying Reasoning Flaws in Planning-Based RL Using Tree Explanations}
\author{
Kin-Ho Lam\and
Zhengxian Lin\and
Jed Irvine\and
Jonathan Dodge\and
Zeyad T Shureih\and
Roli Khanna\and
Minsuk Kahng\and
Alan Fern\and 
\affiliations
}

\begin{document}

\maketitle

\begin{abstract}
Enabling humans to identify potential flaws in an agent's decision making is an important Explainable AI application. We consider identifying such flaws in a planning-based deep reinforcement learning (RL) agent for a complex real-time strategy game. In particular, the agent makes decisions via tree search using a learned model and evaluation function over interpretable states and actions. This gives the potential for humans to identify flaws at the level of reasoning steps in the tree, even if the entire reasoning process is too complex to understand. However, it is unclear whether humans will be able to identify such flaws due to the size and complexity of trees. We describe a user interface and case study, where a small group of AI experts and developers attempt to identify reasoning flaws due to inaccurate agent learning. Overall, the interface allowed the group to identify a number of significant flaws of varying types, demonstrating the promise of this approach. 


\end{abstract}

\section{Introduction}

\label{introduction}
What tools can enable humans to identify flaws in an AI agent's reasoning, even if the agent functions at a super-human level? One possibility is for the AI to produce explanations that simplify decisions into reasoning steps that a human can check against their domain and common sense knowledge. While a human may not be able to fully understand the totality of an AI's decisions, our case study demonstrates flaws in the AI's model can still be identified at the level of reasoning steps. These flaws may suggest ways to improve an agent or warn of potentially serious weaknesses. 
In this paper, we detail a case study exploring how experts can combine common-sense knowledge with a visual state-action tree generated by an AI to seek out flaws in a non-trivial reinforcement-learning (RL) agent.

We consider an agent architecture based on tree search, where each decision is made by constructing a look-ahead search tree that is used to evaluate action choices in the current environment state. The trees serve as an explanation, which breaks down decisions into individual reasoning steps within the tree. These steps include predicting the effects of actions, predicting the values of leaf nodes, and pruning tree branches. Each step can in concept, be inspected by a human to identify potential flaws. 

While tree-based explanations provide the potential for humans to identify step-level flaws, said trees can be large and complex. Thus, it is not clear whether humans will actually be able to detect such flaws in practice given the potentially overwhelming quantity of information.

In this work, we explore this question in the context of a planning-based agent trained to play a complex real-time strategy game (Section \ref{gameRules-information}). We describe the design of an interface for navigating the decision points of game replays and the corresponding trees (Section \ref{sec:interface}). We then describe the experiences of a team of AI experts and developers using the interface to identify flaws (Section \ref{sec:case-study}). We present important design elements of our tree-based explanation interface and how its utility helps experts find flaws in our AI agent. We further present general flaw-finding strategies that may help experts focus their attention to areas likely to contain flaws.

\section{Related Work}
Interestingly, we have not found prior work that provides a principled study of using search trees for explanation purposes. The majority of prior work on explanations for RL concern ``reactive" agents who make decisions via the evaluation of black-box functions. For example, it is common to represent an agent's policy or value function as a neural network, which is simply forward evaluated to make decisions. Explanations, in such cases, may highlight parts of the observations that were important to a decision \cite{pmlr-v80-greydanus18a,DBLP:journals/corr/abs-1809-06061,DBLP:journals/corr/abs-1906-02500,gupta2020explain,atrey2020exploratory}, generate counterfactual inputs \cite{olson2019counterfactual}, extract finite state machines \cite{MISCkoul2018learning}, or display the trade-offs between decisions \cite{juozapaitis2019}. Unfortunately, these types of explanations provide little insight into the internal decision-making process, let alone breaking the decisions into human-consumable steps. As such, these explanations primarily support identifying flaws at the level of full decisions, which may include obvious bad action selection or the agent pays attention to information that is clearly irrelevant. It may be rare, however, for humans to be able to make such judgements for a high-performing AI in a complex environment. 

To support our goals, rather than consider reactive RL agents, we instead focus on planning-based RL agents (Section \ref{learned-components}). 
A planning-based agent learns or is provided a model of the environment. This model is used for decision making via explicit planning. To this end, we designed a stochastic environment that forces the agent to perform long-term planning and decision trade-offs. 

\section{Game Environment: Tug of War}
\label{gameRules-information}

\begin{figure}[htp]
    \centering
    \includegraphics[width=5cm]{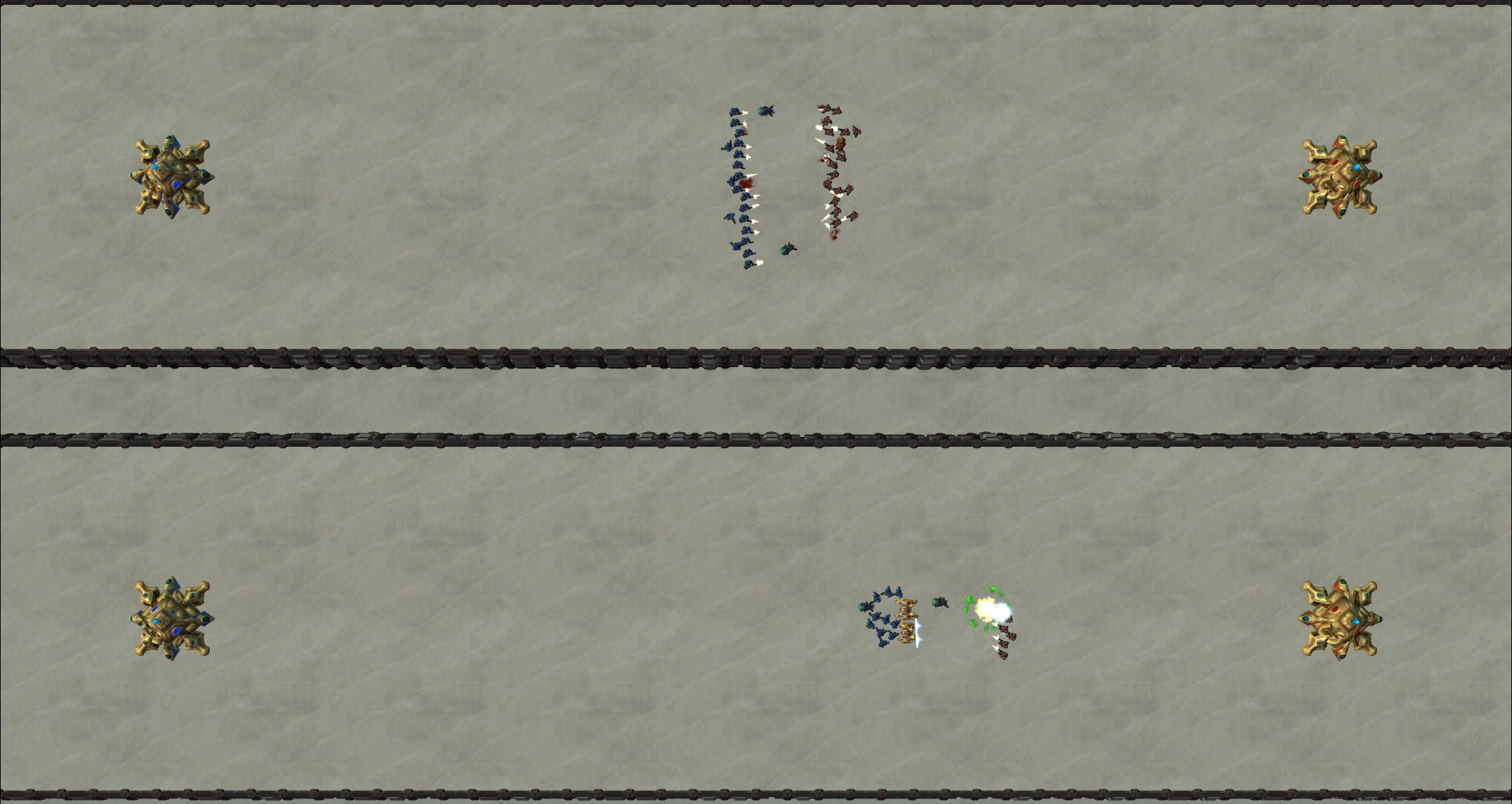} \includegraphics[width=3cm]{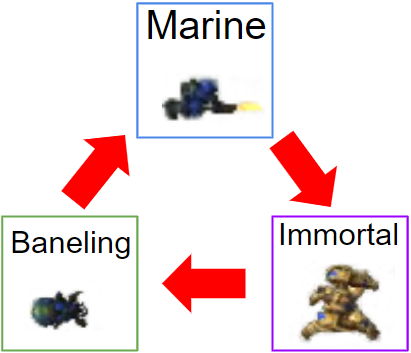}

    \caption{(left) Tug of War game map - Top lane and bottom lane, Player 1 owns the two bases on the left (gold star-shaped buildings), Player 2 owns the two bases on the right. Troops from opposing players automatically march towards their opponent's side of the map and attack the closest enemy in their lane. (right) Unit Rock Paper Scissors - Marines beats Immortals, Immortals beats Banelings, and Banelings beats Marines. We have adjusted unit stats in our custom StarCraft 2 map to befit ToW's balance.}
    \label{fig:ToW-ScreenShot}
\end{figure}

\emph{Tug of War (ToW)} is an adversarial two-player, zero-sum, real-time strategy game our research lab designed for this study using the StarCraft 2 game engine. ToW (Figure \ref{fig:ToW-ScreenShot}) is played on a map divided into top and bottom lanes, where each lane has a two bases (Player 1 - left, Player 2 - right). The game proceeds in 30 second waves, where before the next wave begins each player may select either the top or bottom lane for which to purchase some number of military-unit production buildings (of three different types) with their available currency. These production buildings have different costs. Both Players receive a small amount of currency at the beginning of each wave and a player can linearly increase this stipend by saving to purchase up to three economic buildings, called pylons.


At the beginning of each wave, each building produces one unit of the specified type. The unit automatically walks across the lane to attack incoming enemy units and the opponent's base if close enough. The three unit types, Marines, Immortals, and Banelings, form a rock-paper-scissors relationship (Figure \ref{fig:ToW-ScreenShot}). The first player to destroy one of the opponent bases by inflicting enough damage wins. If no base is destroyed after 40 waves, then the player with the lowest health base loses.

ToW is a challenging for reinforcement learning (RL) due to the large state space, complex dynamics, large action space, and sparse reward. The states give perfect information, although the opponent's unspent currency is not directly revealed (but can be inferred from history). The dynamics have randomness due to variations in damage to buildings and units by attacks, which makes predicting group interactions difficult.
Additionally, the agent's possible actions in any given state can vary widely, from tens to tens of thousands of actions; its options are constrained by the number of ways to allocate its budget.
Finally, the reward is just +1 (winning) or 0 (losing) at the end of a game, which can last for up to 40 waves/decisions. 

\section{Planning-Based RL Agent}
\label{learned-components}

We describe our planning-based RL architecture and learning approach, where decisions are made via tree search using a learned game model and evaluation function. This architecture automatically produces explanations via the search trees, which will be investigated by humans via our explanation interface (Section \ref{sec:interface}). 

{\bf Agent Architecture.} The moment before each new wave is a decision point for the RL agent, where it selects an action based on an abstract, human-understandable, state description. The abstract state includes information about the wave number, health of all bases, the agent's unspent currency, both player's current building counts, pylon quantities, and the number of friendly and enemy troops of each type in each of 4 evenly divided grid cells per lane. The main form of abstraction is the positions of units, which are quantized into 8 grid cells and health of individual units is not provided. Nevertheless, this abstraction is rich enough for humans to understand the state and make reasonable decisions. We will also see it is rich enough to allow for strong AI performance.

Our agent architecture is similar to AlphaZero \cite{silver2018}, an RL agent based on game-tree search that demonstrated mastery of Go, Chess, and Shogi. 
Unlike AlphaZero, however, our agent uses a learned model of the game dynamics. The agent is constructed from the following components: 1) \emph{Learned Transition Model}, which predicts the next abstract state given a current abstract state and actions for both players. Note that the next state (next wave) occurs 30 seconds after the current state. 2) \emph{Learned Action Ranker}, which returns a numeric ranking over actions in an abstract state based on their estimated quality. 3) \emph{Learned State-Value Function}, which returns a value estimate (probability of winning) given an abstract state. 

Given the learned components, the agent selects a decision point by building a pruned depth limited game tree. Specifically, the search is parameterized by the search depth $D$ and for each depth $d\in \{0,\ldots, D-1\}$ the number of friendly $f_d$ and enemy $e_d$ actions to be expanded. Starting at a node corresponding to the current state, the search uses the action ranker to expand the appropriate number of friendly and enemy actions, and for each combination uses the transition model to produce the predicted state at the next depth level. The tree is expanded until reaching nodes at depth $D$, which are assigned values by the state-evaluation function. The minimax value of the root actions are computed and the best action is selected. In our case study, we use $D=2$ (i.e. 1 minute of game time) with $f_d = (20, 5)$ and $e_d = (10, 3)$. Thus, the root node selects among 20 possible actions in the current state. Figure \ref{fig:ToW-ReplayAndExpl} gives a visualization of a game tree from a particular state.



{\bf Learning Approach.} To learn the required components we use model-free RL to learn a Q-function $Q(s,a)$ that estimates the value of action $a$ in abstract state $s$. Specifically, since this is a two-player game, we use a tournament-style self-play strategy, where a pool of previously trained model-free agents, each with their own Q-function, is used to play against a currently learning agent. The current agent is trained until it achieves a high win-percentage against the pool or training resources are expended. This is similar to the self-play strategy employed by AlphaStar for the full game of StarCraft 2 \cite{vinyals2019}. 

To train each agent we use a variant of DQN \cite{mnih2015} called Decomposed Reward DQN \cite{juozapaitis2019}, allowing us to learn a Q-function that returns a vector of probabilities over the different win conditions for each player. The sum of the probabilities equals 1 and the sum over win conditions for a player is the is the value of the action. This vector provides more insight into the agent's decision making and part of the visualization in our explanation interface (Section \ref{sec:interface}). The Q-function of the best agent in the pool (typically the last agent) is used as the learned action ranking for search. In addition, it is also used for the state-evaluation function by letting the state value to be the value of the best action. 

We represent the Q-function using a 3 layer feed-forward neural network with an input consisting of features describing the abstract state and action. This outputs a value vector of the state-action pair. Self-play training was conducted for two days, after which the learned agent appeared to be quite strong, likely better than most humans with some game experience.

To learn the dynamics model used for search, we used two sources of data: 1) Data from games between agents in the tournament and random agents, 2) The replay buffer of the final learned agent, which is accumulated as part of the learning process. Each data instance was of the form $(s, a_f, a_e, s', \vec{r})$ giving the current abstract state, friendly action, enemy action, next abstract state, and decomposed-reward vector respectively. Here the reward vector is the zero vector for all states, except at the end of the game where it is the zero vector for a loss and a one-hot encoding of the win condition otherwise. We designed a feed-forward neural network that takes $s$, $a_f$, and $a_e$ as input to predict $s'$ and $\vec{r}$. Note that this approximates the dynamics as a deterministic function. While the actual dynamics are stochastic due to unit level randomization of damage, in aggregate, a deterministic model appears to be adequate for strong play. 


\section{Explanation Interface}
\label{sec:interface}

The above planning-based RL agent produces a search tree that serves as an explanation of its decisions at each decision point. While this is a sound certificate of the decision, it may be difficult or impossible for a human to ``fully understand" the reasoning steps leading to the overall decision. This is especially true when the agent is playing at a superhuman level. Rather, a human can still consider the individual learned inferences involved in the search to build confidence in the agent. If the agent perfectly learns the search knowledge (model and value functions), the agent will play perfectly. Thus, any sub-optimal play is due to inaccuracies in the learned components. Our explanation interface is designed to allow humans to explore tree explanations in order to identify such inaccuracies, which can then be used to build confidence or improve the agent. 

\begin{figure}[tp]
    \centering
    \includegraphics[width=6cm]{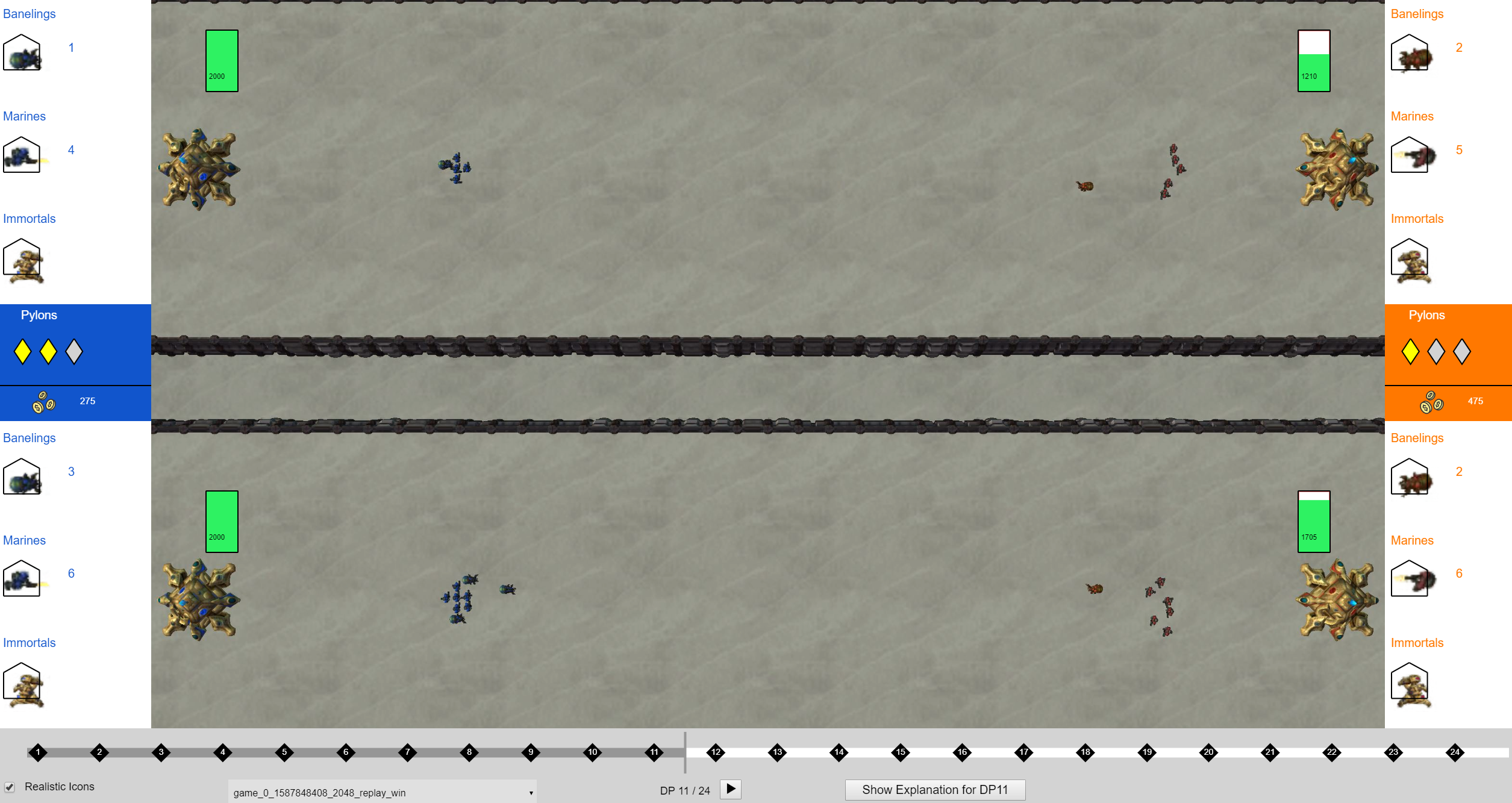}
    \caption{ToW Replay Interface - Displays a visual replay of the ToW game, including unit building quantities for each lane, currency, pylons, base health, and the player's actions as they happen.}
    \label{fig:ToW-tree-Ui}
\end{figure}

{\bf Replay Interface.} Figure \ref{fig:ToW-tree-Ui} illustrates the ToW replay interface, which allows a user to navigate games played by the AI. The user can let the game play, pause, or jump to a specific numbered decision point by clicking on the timeline at the bottom of the screen. After selecting a decision point, a high-level view of the agent's choice is shown via a graph of the win-probability estimates computed by the tree for the 20 root actions in sorted order (see Figure \ref{fig:ToW-ScreenShot}). This visualization is used to help select interesting decision points to investigate (Section \ref{sec:case-study}). At any decision point the user can click on the ``Show Explanation" button to bring up the tree explanation interface. 

%
%

\begin{figure*}
    \centering
    \includegraphics[scale=0.3]{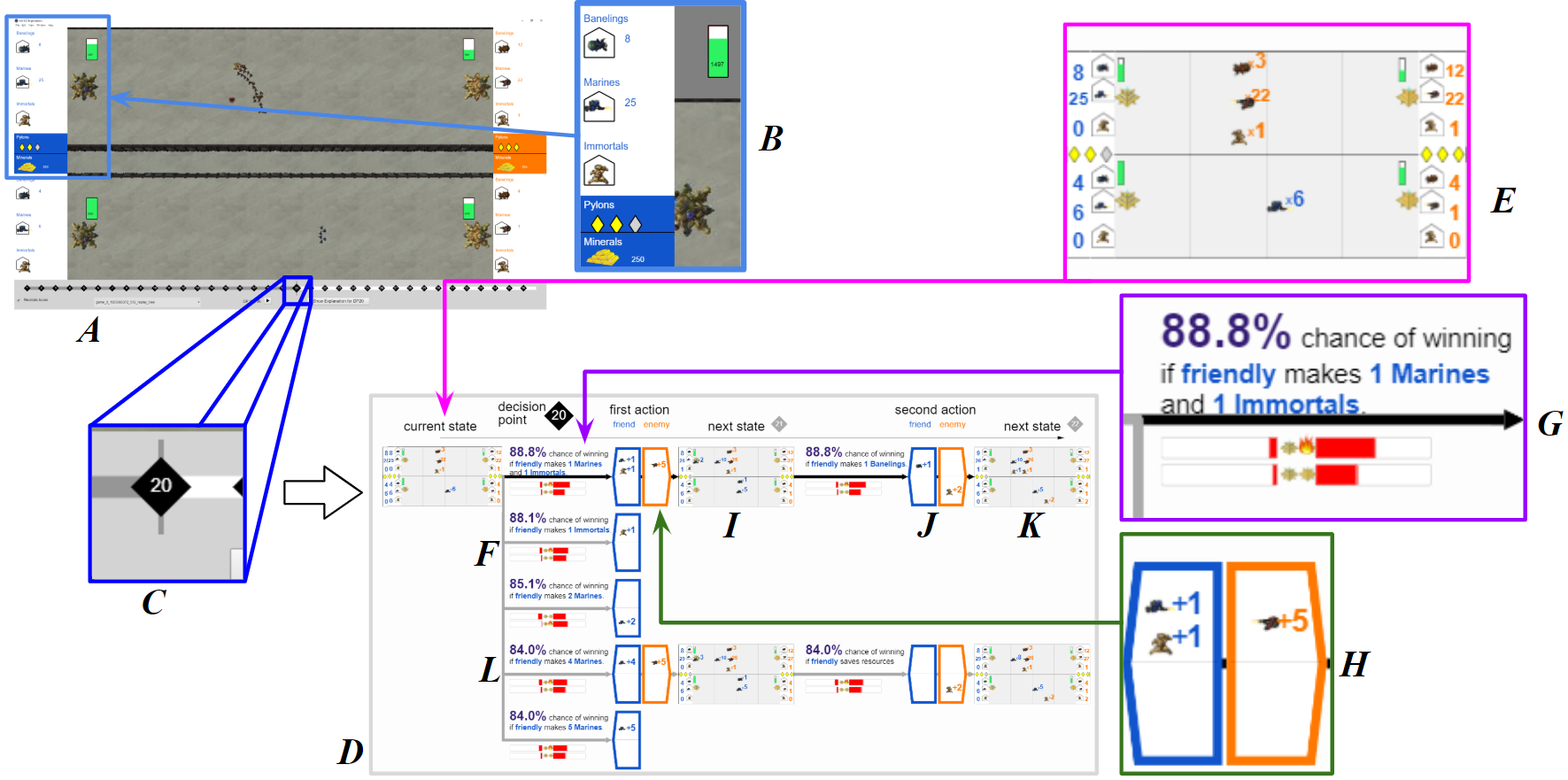}
    \caption{
    Tug of War Explanation interface, replay interface is paused at D20 and D20's explanation tree is shown.
    \textbf{\textit{A}}: ToW Replay Interface paused at D20.
    \textbf{\textit{B}}: ToW Replay Interface Friendly Top unit building count and base health.
    \textbf{\textit{C}}: Replay player is paused at D20.
    \textbf{\textit{D}}: Tree interface at D20. Top 5 actions from the current state shown. Topmost action (principal variation) and 4th best action are fully expanded.
    \textbf{\textit{E}}: Game state at D20 as shown in the replay interface. This image is generated using the model-based agent's game-state input vector.
    \textbf{\textit{F}}: Other possible lower ranked actions are shown below. Evaluator can expand or hide action nodes to compare futures.
    \textbf{\textit{G}}: 88.8\% expectation of winning by destroying P2's Top Base at D22.
    \textbf{\textit{H}}: Action node, the agent expects the opponent's best action is to purchase +5 marine buildings in the top lane. Using min-max approximation, the agent believes it should build +1 marine buildings and +1 immortal buildings in the top lane.
    \textbf{\textit{I}}: Expected (Predicted) state D21 if both players perform action node \textit{H}.
    \textbf{\textit{J}}: Action node, best min-max action to produce optimum state \textit{K} from D21 to D22. 
    \textbf{\textit{K}}: Expected state at D22 if min-max action pairs \textit{H} and \textit{J} are taken.
    \textbf{\textit{L}}: 4th best action with an expected 84.0\% chance of a favorable D22 state. Path has been expanded to show a possible alternative future for D22 if different actions are taken.
  }
  \label{fig:ToW-ReplayAndExpl}
\end{figure*}

{\bf Tree Explanation Interface.} The tree explanation interface allows a user to visually analyze parts of the agent's search tree at a decision point. The interface shown in Figure \ref{fig:ToW-ReplayAndExpl}, illustrates part of a tree. Figure \ref{fig:ToW-ReplayAndExpl}D is associated with a selected decision point (Figure \ref{fig:ToW-ReplayAndExpl}C). The tree is visualized as left-to-right paths from the current root state (left-most) to leaf nodes. Paths are displayed via visualizations of abstract state nodes, followed by visualizations of actions, which transition to predicted abstract state nodes. 

Figure \ref{fig:ToW-ReplayAndExpl}{E} shows an abstract state node; this node visually displays the abstraction data used by the agent to represent states. Base damage prediction in future state nodes are illustrated by red and green base health bars as shown in Figure \ref{fig:HPError}. Red indicates damage inflicted and green indicates remaining health. The house-shaped boxes (Figure \ref{fig:ToW-ReplayAndExpl}H) contain the actions performed by each player from the parent state node. Notice the houses are divided horizontally by a faint grey line thereby distinguishing whether the action was performed in the top or bottom lane.

Initially the user is only shown the principle variation path (i.e. the best mini-max path). From that point, the user can expand to the next-best action at a node, or expand a path to the maximum depth along the principle variation from that node. The interface also allows the user to rearrange paths for easier comparison. For each expanded action we depict the game outcome probabilities (Figure \ref{fig:ToW-ReplayAndExpl}G), which show the outcome expected by the agent for each action based on the tree. The outcome information on actions leading to leaf nodes is directly computed by the learned state-evaluation function. Using this interface, a user can explore the tree to identify potential flaws in the learned knowledge, which might correspond to flaws in the predicted state transitions or flaws in the state evaluation function.

\section{Case Study}
\label{sec:case-study}

This case study illustrates how AI experts and developers can use tree explanations to identify flaws in the underlying learned components of an RL agent. The users of the system included 3 of the main system software developers and 2 of the primary designers of the agent architecture and learning algorithms. These individuals worked together over multiple sessions totalling approximately 4 hours to identify as many flaws as possible. Note that while much explainable AI work is targeted toward non-experts in AI, here we focus on an expert population. 
Such experts are likely to be the first users who benefit from such tools. Further, it is reasonable to first demonstrate that experts can effectively use an explainable AI tool before investigating whenever non-AI experts can use the tools effectively.
%
Below, we first describe the strategies employed to identify likely decision points that contain flaws. Next we present several examples of flaws found during the investigation.  





{\bf Selecting Games and Decision Points.}
\label{analysisSelection}
It is important to have strategies for zeroing in on decision points that are likely to uncover flaws. While this is a process that future versions of the explanation system might help with, it is currently a largely a manual process based on intuition. 
An important are of future work is to develop effective automated heuristics for the selection of promising states to investigate. For example, other researchers have proposed specific metrics for ``criticality'' of states based on the difference between the maximum Q-value and average Q-value at a state---\emph{``those in which acting randomly will produce a much worse result than acting optimally''}~\cite{Huang2018}.

We first chose to focus our evaluation on games the agent lost because one can anticipate reasoning errors to have contributed to the agent losing.
Thus, we collected a set of 13 losing games by having the agent play a large number of games against other agents in the tournament pool.
We quickly discovered that looking at early decision points was not productive because most actions had uniform win probabilities and the states had little information. This makes sense since early in a game, even bad action choices can be recovered from by a strong agent, which results in similar values across the entire action set. 

\begin{figure}[tp]
    \centering
    \includegraphics[width=4cm]{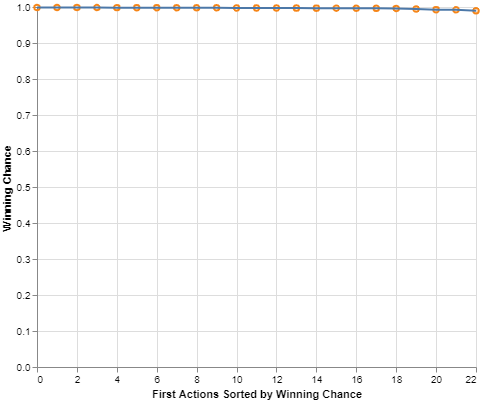}
    \includegraphics[width=4cm]{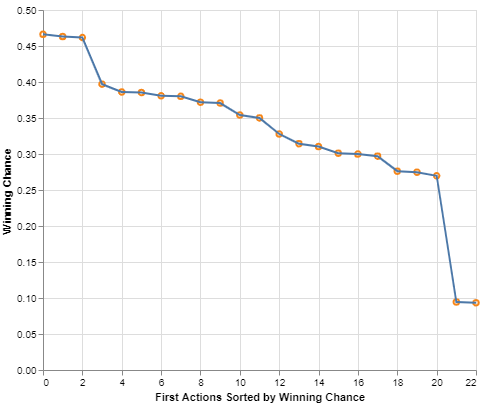}
    \caption{(left) Optimistic Win Expectation - The model based agent expects 2 states into the future, any of the top 22 actions it has evaluated will have a 100\% chance of winning the game. (right) Model Based Agent Large Uncertainty - An example of a decision point with relatively low win expectations that drop sharply. We hypothesize decision points with huge drops in win expectation contain clear flaws.  }
    \label{fig:ToW-q-Chart}
\end{figure}

Thus, the team used the interface to focus on two kinds of states: just prior to sudden drops in win probability estimates (indicating a possible mistake just occurred) and where the action probabilities showed visibly apparent fluctuation across the actions, as shown in Figure~\ref{fig:ToW-q-Chart}.
The team identified candidate states using the above criterion and then investigated the corresponding games around those points. The following describes some example flaws that were identified in this way. 

\begin{figure}[t]
    \centering
    \includegraphics[width=8.5cm]{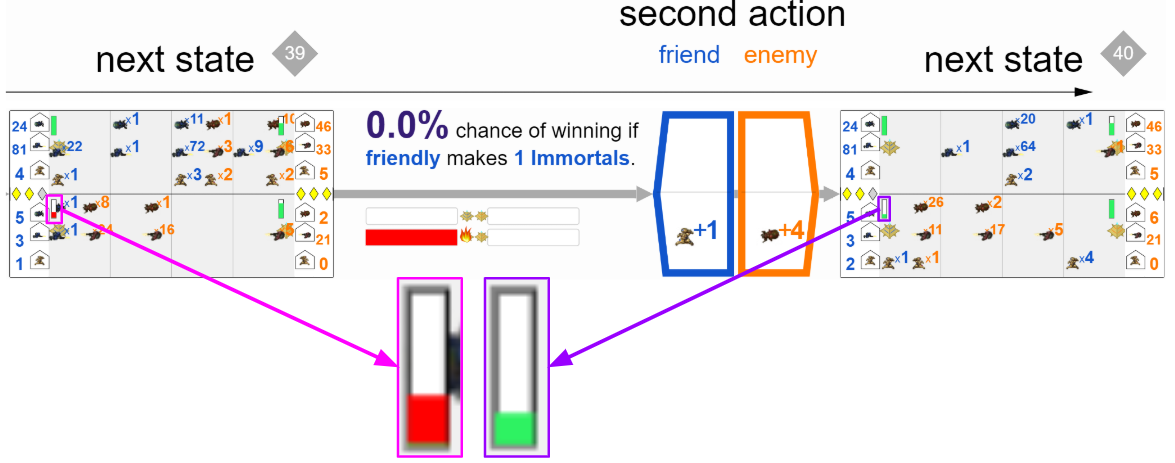}
    \caption{Base Health Prediction Error; Red health bars indicates expected damage inflicted, green indicates expected remaining health. In this case, the agent expects to have more base health in D40 than in D39, which is not possible.}
    \label{fig:HPError}
\end{figure}

{\bf Transition Function Flaws.} Recall Tug of War is an adversarial two-player game where the objective is to inflict the most damage on the opponent’s base. Damage is dealt when units reach attacking range of the enemy’s base. It is a fundamental game rule that damage is irreversible; the health of a base cannot increase. When investigating the tree resulting from a suspicious decision point, which led to a sudden drop in win probability, we identified an error in the state transition function, where the health of a base increased (Figure \ref{fig:HPError}). Here, at decision point 39, Player 1's bottom base is predicted to sustain a significant amount of damage and will have almost no health remaining. However at D40 Player 1's bottom base is predicted to have more health than in D39. This is a serious error that clearly demonstrates the AI has not yet fully learned some key constraints of the game. Such flaws can lead to rare, but critical mistakes in evaluating actions. The flaw suggests that more training is required in situations where the agent's bases have very low health, which was relatively rare late in training as the agent frequently wins. 

\begin{figure}[t]
    \centering
    \includegraphics[width=8.5cm]{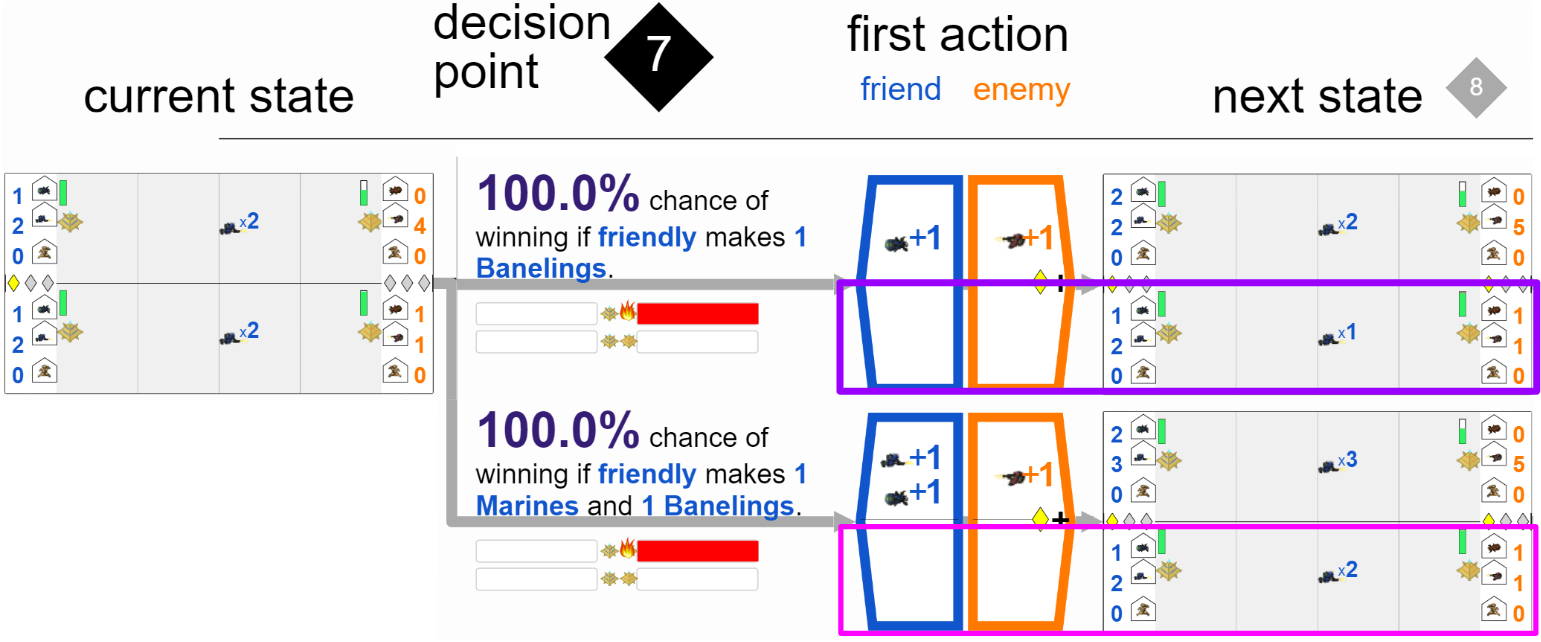}
    \caption{Lane independence prediction violation; No action was taken in the bottom lane by either player, yet different quantities of marines are expected to be on the field in the bottom lane despite no change.}
    \label{fig:IndependenceError}
\end{figure}

Recall at each decision point, the agent is only able to purchase buildings in either the top or bottom lane. If both players happen to choose the same lane, the other lane will not be affected. Figure \ref{fig:IndependenceError}, illustrates a situation where two actions from the root state have both players selecting the top lane. This implies the bottom lane for the predicted states of those actions should be identical. However, we see this is not the case; instead the number of predicted marines in the bottom lane differs for the two states. This shows that the transition function has not yet sufficiently learned this basic lane-independence property. Again, this suggests more training is necessary, which might include specialized training data augmentation that captures the invariance. 


\begin{figure}[t]
    \centering
    \includegraphics[width=8.5cm]{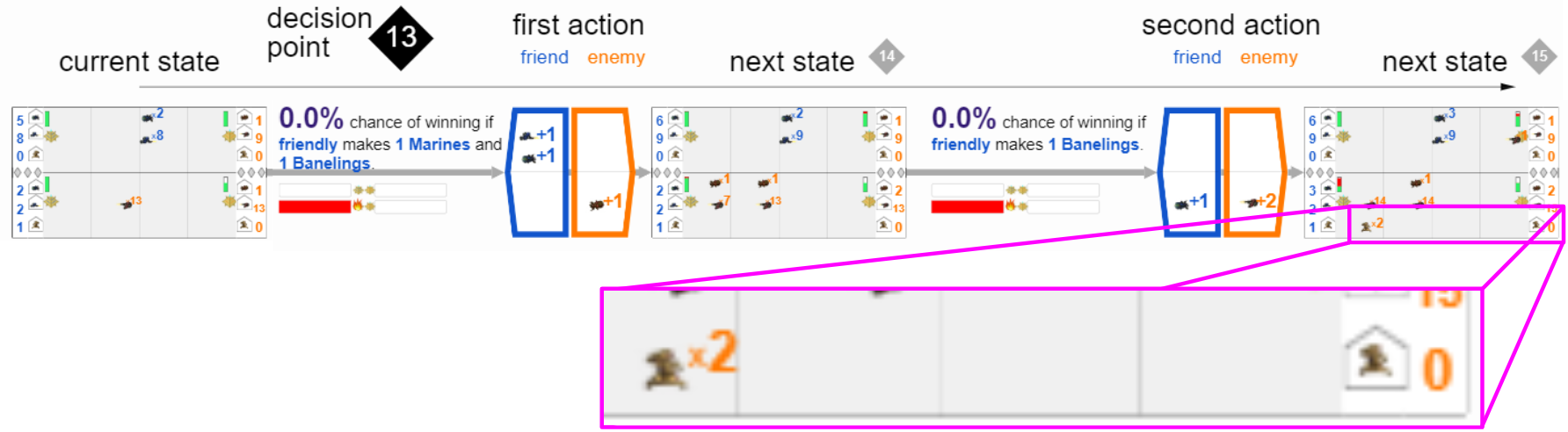}
    \caption{Transition Function Error: P1 expects opponent P2 to have two immortal units in the bottom lane even though P2 lacks the necessary production buildings. This predicted state in D15 is not possible.}
    \label{fig:unitsNoBUildings}
\end{figure}

Figure \ref{fig:unitsNoBUildings} depicts player 2 has two immortal units on the field, even though player 2 does not have any buildings to produce immortals. This is an impossible state, and this type of error could completely change the evaluation of a state and its upstream root actions. The team hypothesizes the agent hasn't observed enough situations with immortals in the bottom lane as the agent (P1) rarely uses immortals there. We also notice P1 has an immortal building in the bottom lane. It is possible the agent has not encountered enough situations with immortals in the bottom lane and has not learned the unit/building association. 

\begin{figure}[t]
    \centering
    \includegraphics[width=8.5cm]{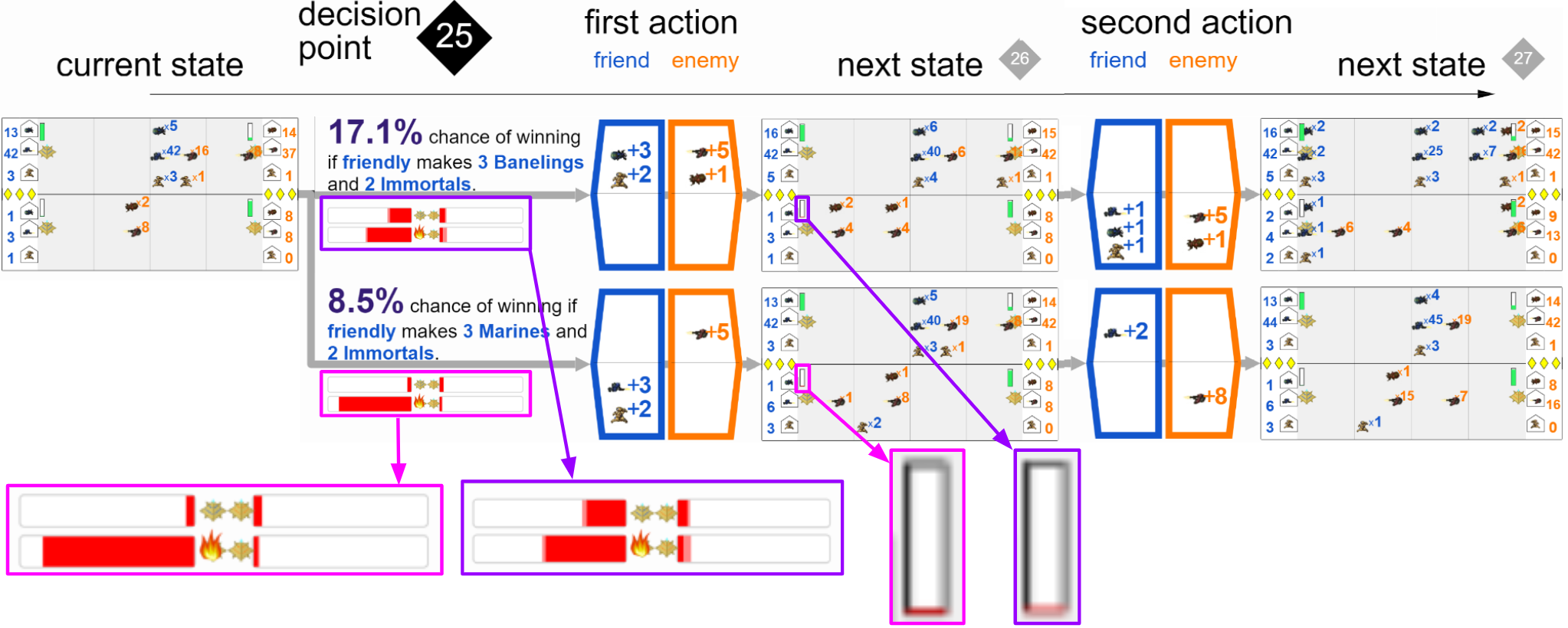}
    \caption{Terminal State Ranking Error; Player 1 predicts two futures where it expects it will lose in D26. However the probability that it will win the game is non-zero in both cases because it has appended D27, a state that cannot exist because the previous state would have ended the game.}
    \label{fig:TerminalStateRanking}
\end{figure}

{\bf State Evaluation Function Flaws.} Figure \ref{fig:TerminalStateRanking} depicts predicted paths where the agent fails to recognize it expects to have lost the game in state D26, thus incorrectly ranking the whole path. As shown in Fig. \ref{fig:TerminalStateRanking}, the agent expects its bottom base to get destroyed, its bottom base has approximately 0 health at D26. However, its Transition function predicts a game state for D27 and the state evaluation function assigns it a non-zero probability of winning. This results in the friendly agent enacting a move on an impossible state, and the value of the impossible state is recursed back to the root action resulting in a state with an expected losing future getting ranked higher than an alternative state with an arguably better move. This flaw might also be considered to be a flaw in the tree construction procedure, where the predicted health should have been considered to be zero which halts tree construction. 

\begin{figure}[tp]
    \centering
    \includegraphics[width=8.5cm]{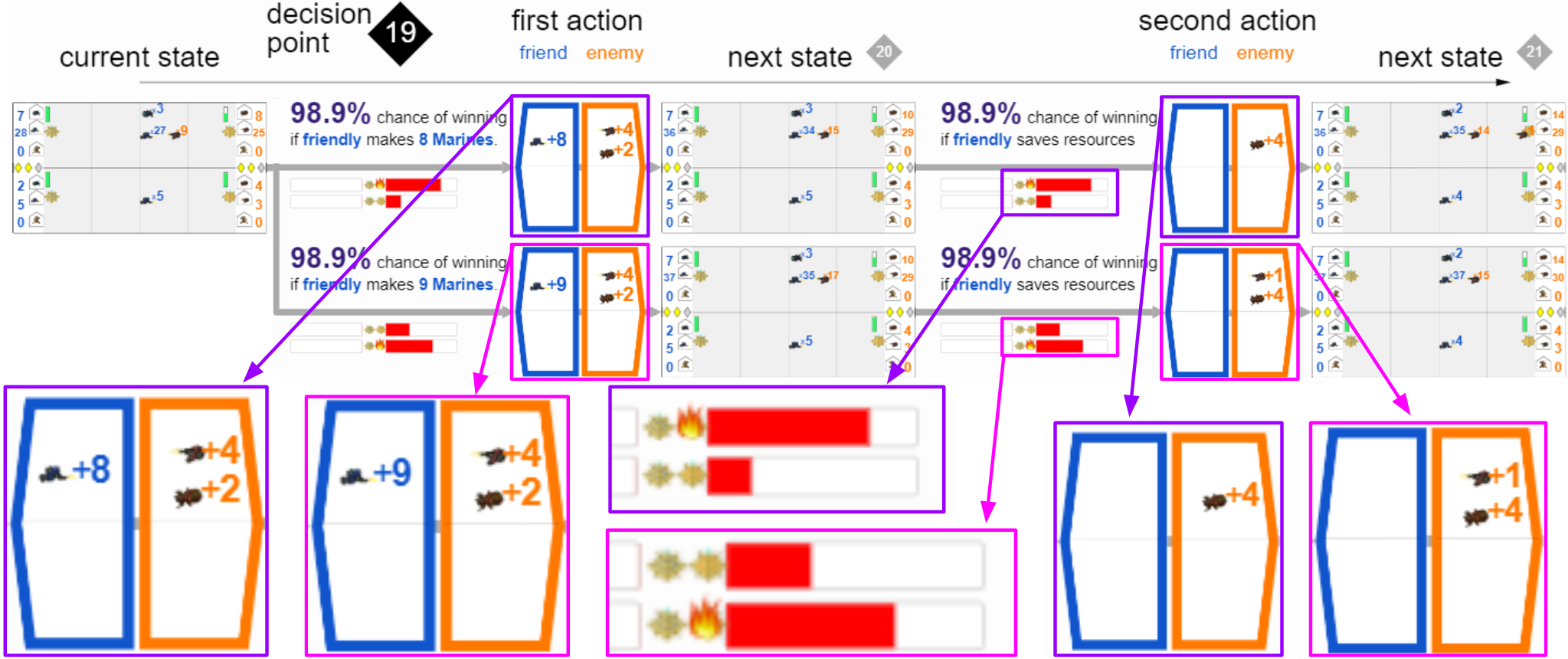}
    \caption{State Evalution Error; Marginal difference in actions, yet win expectation flips from destroying the enemy's top base to destroying the enemy's bottom base.}
    \label{fig:winSwap}
\end{figure}

As another example, in Figure \ref{fig:winSwap}, we see two very similar lines of action from the root state, that result in qualitatively very different state evaluations. In particular, the lower action sequence leads to a leaf state where the probability of destroying the bottom base is high. However, the upper action sequence leads to a very similar leaf state, but the probability of destroying the top base is high. The similarity of these action sequences and leaves, combined with the differences in the predicted outcome goes against common sense. This highlights an apparent weakness in the learned state-evaluation function, which could potentially result in inaccurate action selections under the right conditions. 


{\bf Scripting Flaw Detection.} One aspect of the above examples is that in most cases, it was possible to describe the type of flaw in a simple and general way. For example, the transition function incorrectly predicted the increase in a base's health. This suggests that human-guided flaw finding can be used to then automatically check for other similar flaws automatically. In particular, one can write a script that checks a library of games for a general description of a perceived flaw to determine how wide-spread it is. 

As an example of this the team wrote a script to search for the type of health increase flaws observed in Figure \ref{fig:HPError}. The script searched all states in all trees for all decision points in 6 losing games played by our agent. It was discovered the agent expected base health to rise in the decision point preceding the moment before losing in 4 out of 6 losing games. Our scripts also indicated the agent made the base health misjudgement by a large margin in some cases. For example, there were instances where it expected the health to rise by more than 10\%, which could easily lead to misjudements in close games. 

We expect this strategy of using such an explanation interface to extract general descriptions of potential flaws will be a powerful tool for validating and improving complex AI systems. The results of running the scripts can help direct AI engineers to address the most pressing flaws given limited resources.

\section{Case Study Summary}  

Our case study demonstrated it was possible for a team of expert AI researchers and developers to identify non-trivial bugs using a relatively simple explanation interface. Although reasoning errors were not necessarily visible in the overall action selected by the agent, the tree interface allows for underlying flaws to be identified. This demonstrates the utility of this type of ``micro decision" analysis of an agent's decision making. Our study also demonstrated the flaws that were found tended to have simple and general descriptions and suggested those descriptions can be used to search for other similar flaws. 


As noted earlier, the model-based agent receives a coarse abstraction of the ToW game state and can only provide tree explanations in terms of this  abstraction. Given how explanation trees can easily span thousands of considered futures, with each state node communicating a dense set of information, organizing and presenting said information is a challenge. The size of the data presented also draws human comprehension/attention into question, a structured evaluation process may be needed to focus manual review to critical sections where humans follow a procedure to dissect a tree; else an unguided process resembles searching for a needle in a haystack. In this case study, the team was successful in identifying promising points of investigation, but in general that may not be the case. 

While agents are able to converge to a policy after playing millions of ToW games, designing an interface to communicate the game state from the agent's perspective is no trivial task. Our explanation interface design for Tug of War continues to evolve as we consider different audiences with varying levels of AI expertise.


\section{Future Work}



One of the key enablers of our case study's explanation interface is the abstract state representation used by the agent; humans can interpret the tree data. As environments get ever more complex and closer to perceptual data, the internal representations of the agent will have less correspondence with what can be visualized by a human. This is particularly true in systems where the underlying state space used for search is a learned latent space with no presupposed semantics. Thus, a key direction for future research is to identify ways to make the correspondence as tight as possible without sacrificing agent performance. Another key direction of future work is to assist users in identifying promising areas of investigation and to provide a measure of ``testedness" that indicates how similar parts of the tree are to previously ``verified" parts of trees. This would provide users with a sense of progress and improve the efficiency of flaw finding. Finally, flaw finding is just one goal of explainable AI. It is worth considering other objective goals and understanding the types of interfaces and processes that best suit each one. 

\section{Acknowledgements}
This work is partially supported by DARPA under the grant N66001-17-2-4030.

\bibliographystyle{plain}
\bibliography{paper.bib}

\end{document}